\documentclass[runningheads]{llncs}
\usepackage{amsmath,amssymb}
\usepackage{algorithmic}
\usepackage[flushleft]{threeparttable}
\usepackage{float}
\usepackage{multirow,makecell,multicol}
\usepackage{graphicx}
\usepackage[breaklinks=true,letterpaper=true,colorlinks,bookmarks=false]{hyperref}
\usepackage{array}
\usepackage[table]{xcolor}
\usepackage{xcolor}
\def\figurename{Fig.}
\def\sectionname{Sec.}
\def\tablename{Table}
\newcolumntype{P}[1]{>{\centering\arraybackslash}p{#1}}
\definecolor{maroon}{RGB}{58, 183, 149}
\definecolor{iblue}{RGB}{33, 53, 236 }
\definecolor{tablered}{cmyk}{0,0.87,0.68,0.32}

\usepackage{soul}
\newcommand{\etal}{\mbox{\emph{et al.\ }}}
\newcommand{\ie}{\mbox{\emph{i.e.,\ }}}
\newcommand{\eg}{\mbox{\emph{e.g.,\ }}}
\definecolor{newblue}{RGB}{23, 107, 239}

\PassOptionsToPackage{hyphens}{url}\usepackage{hyperref}

\begin{document}
\title{A Systematic Benchmarking Analysis of Transfer Learning for Medical Image Analysis}
\titlerunning{Benchmarking Transfer Learning for Medical Image Analysis}
%
 \author{Mohammad Reza Hosseinzadeh Taher\inst{1} \and
Fatemeh Haghighi\inst{1} \and
Ruibin Feng\inst{2} \and
Michael B. Gotway\inst{3} \and
Jianming Liang\inst{1}}
 

\authorrunning{MR. Hosseinzadeh Taher et al. }
%
\institute{Arizona State University, Tempe, AZ 85281, USA 
\email{\{mhossei2,fhaghigh,jianming.liang\}@asu.edu} \and
Stanford University, Stanford, California 94305, USA \\
\email{ruibin@stanford.edu} \and
 Mayo Clinic, Scottsdale, AZ 85259, USA\\
\email{Gotway.Michael@mayo.edu}}

\maketitle              
\begin{abstract}
   Transfer learning from supervised ImageNet models has been frequently used in medical image analysis. Yet, no large-scale evaluation has been conducted to benchmark the efficacy of newly-developed pre-training techniques for medical image analysis, leaving several important questions unanswered. As the first step in this direction, we conduct a systematic study on the transferability of models pre-trained on iNat2021, the most recent large-scale fine-grained dataset, and 14 top self-supervised ImageNet models on 7 diverse medical tasks in comparison with the supervised ImageNet model. Furthermore, we present a practical approach to bridge the domain gap between natural and medical images by continually (pre-)training supervised ImageNet models on medical images. Our comprehensive evaluation yields new insights:  (1) pre-trained models on fine-grained data yield distinctive local representations that are more suitable for medical segmentation tasks, (2) self-supervised ImageNet models learn holistic features more effectively than supervised ImageNet models, and (3) continual pre-training can bridge the domain gap between natural and medical images. We hope that this large-scale open evaluation of transfer learning can direct the future research of deep learning for medical imaging. As open science, all codes and pre-trained models are available on our GitHub page \url{https://github.com/JLiangLab/BenchmarkTransferLearning}.

\end{abstract}

\section{Introduction}
\label{sec:intro}
To circumvent the challenge of annotation dearth in medical imaging, fine-tuning supervised ImageNet models (i.e., models trained on ImageNet via supervised learning with the human labels) has become the standard practice~\cite{haghighi2021transferable,tajbakhsh2016convolutional,shin2016deep,ZHOU2021Models,haghighi2020learning}. As evidenced by~\cite{ZHOU2021Models}, nearly all top-performing models in a wide range of representative medical applications, including classifying the common thoracic diseases, detecting pulmonary embolism (PE), identifying skin cancer, and detecting Alzheimer’s Disease, are fine-tuned from supervised ImageNet models.
However, intuitively, achieving outstanding performance on medical image classification and segmentation would require fine-grained features. For instance, all chest radiographs (CXR)  have a relatively similar appearance;  therefore,  identifying abnormal conditions and diagnosing specific disorders often rely on recognition of subtle image details.
Furthermore, delineating organs and isolating lesions in medical images would demand some fine-detailed features to determine the boundary pixels. In contrast to ImageNet, which was created for coarse-grained object classification, iNat2021~\cite{vanhorn2021benchmarking}, the most recent large-scale fine-grained dataset, has recently been created. It consists of 2.7M training images covering 10K species spanning the entire tree of life. As such, the {\bf first question} this paper seeks to answer is: 
{\em What advantages can supervised iNat2021 models offer for medical imaging in comparison with supervised ImageNet models?}

In the meantime, numerous self-supervised learning (SSL) methods have been developed. In the afore-discussed transfer learning, models are pre-trained in a supervised manner using expert-provided labels. By comparison, SSL pre-trained models use machine-generated labels. 
The recent advancement in SSL has resulted in self-supervised pre-training techniques that surpass gold standard supervised ImageNet models in a number of computer vision tasks~\cite{wei2020semantic,ericsson2021selfsupervised,islam2021broad,zhao2021makes,caron2021unsupervised}. Therefore, the {\bf second question} this paper seeks to answer is: {\em How generalizable are the self-supervised ImageNet models to medical imaging in comparison with supervised ImageNet models?}

More importantly, there are significant differences between natural and medical images. Medical images are typically monochromatic and typically contain consistent anatomical structures~\cite{haghighi2021transferable,haghighi2020learning}. Recently, several moderately sized medical imaging datasets have been created, including NIH ChestX-Ray14~\cite{wang2017chestx}, which contains 112K images, and CheXpert~\cite{irvin2019chexpert}, which contains 224K images. Naturally, the {\bf third question} this paper seeks to answer is: {\em Can these moderately-sized medical image datasets help bridge the domain gap between natural and medical images?}

To answer these questions, we conduct the first extensive benchmarking study to evaluate the efficacy of different pre-training techniques for diverse medical imaging tasks, covering various diseases (e.g., PE, pulmonary nodules, tuberculosis, etc), organs (e.g., lung and optic fundus), and modalities (e.g., CT, X-ray, and funduscopy). Concretely, (1) we study the impact of pre-training data granularity on transfer learning performance by evaluating the fine-grained pre-trained models on iNat2021 for various medical tasks; (2) we evaluate the transferability of  14 state-of-the-art self-supervised ImageNet models to a diverse set of tasks in medical image classification and segmentation; and (3) we investigate domain-adaptive (continual) pre-training~\cite{gururangan2020dont} on natural and medical datasets to tailor ImageNet models for target tasks on chest X-rays. 

Our extensive empirical study reveals the following important insights:
(1) Pre-trained models on fine-grained data yield distinctive local representations that are beneficial for medical segmentation tasks, while pre-trained models on coarser-grained data yield high-level features that prevail in classification target tasks (see \figurename~\ref{fig:imagenet_vs_inat}).
(2) For each target task, in terms of the mean performance, there exist at least three self-supervised ImageNet models that outperform the supervised ImageNet model, an observation that is very encouraging, as migrating from conventional supervised learning to self-supervised learning will dramatically reduce annotation efforts (see \figurename~\ref{fig:imagenet_vs_ssl}). 
(3) Continual (pre-)training of supervised ImageNet models on medical images can bridge the  gap  between  the  natural and  medical domains, providing more powerful pre-trained models for medical tasks (see \tablename~\ref{tab:seq_pretraining}). 

\section{Transfer Learning Setup}
\noindent\textbf{Tasks and datasets:} \tablename~\ref{tab:datasets} summarizes the tasks and datasets, with more additional details provided in Appendix~\ref{sec:appendix_datasets}. We considered a diverse suite of 7 common but challenging medical imaging tasks encompassing various diseases, organs, and modalities. These tasks span many common properties of medical imaging tasks, such as imbalanced classes, limited data, and small-scanning areas for pathology of interest. We use official data split of these datasets if available; otherwise, we randomly divide the data into 80\%/20\% for training/testing.

\begin{table*}[t]
\begin{center}
\begin{threeparttable}
\caption{
Benchmarking transfer learning for seven common medical imaging tasks, spanning over different label structures (binary/multi-label classification and segmentation), modalities, organs, diseases,  and data size.
}
\label{tab:datasets}
\begin{tabular*}{\textwidth}
{p{0.085\linewidth}p{0.452\linewidth} p{0.15\linewidth}p{0.32\linewidth}}
\hline
Code$^{\dagger}$ & Application  & Modality & Dataset \\
\hline
\texttt{ECC}& Pulmonary Embolism Detection & CT & RSNA PE Detection~\cite{PEchallenge}  \\
\texttt{DXC$_{14}$}& Fourteen thorax diseases classification  & X-ray & NIH ChestX-Ray14~\cite{wang2017chestx} \\
\texttt{DXC$_{5}$}& Five thorax diseases classification  & X-ray & CheXpert~\cite{irvin2019chexpert} \\
\texttt{VFS}& Blood Vessels Segmentation  & Fundoscopic & DRIVE~\cite{Budai2013Robust} \\
\texttt{PXS}& Pneumothorax Segmentation &X-ray& SIIM-ACR~\cite{PNEchallenge}\\
\texttt{LXS}& Lung Segmentation & X-ray&  NIH Montgomery~\cite{Jaeger2014Tow} \\
\texttt{TXC}& Tuberculosis Detection & X-ray& NIH Shenzhen CXR~\cite{Jaeger2014Tow} \\
\hline
\end{tabular*}

\begin{tablenotes}
        \item $^{\dagger}$ The first letter denotes the object of interest (``\texttt{E}'' for embolism, ``\texttt{D}'' for thorax diseases, etc); the second letter denotes the modality (``\texttt{X}'' for X-ray, ``\texttt{F}'' for Fundoscopic, etc);  the last letter denotes the task (``\texttt{C}'' for classification, ``\texttt{S}'' for segmentation).
\end{tablenotes}
\end{threeparttable}
\end{center}
\end{table*}

\medskip
\noindent\textbf{Evaluations:} 
We evaluate various models pre-trained with different methods and datasets. Therefore, we control other influencing factors such as preprocessing, network architecture, and transfer hyperparameters.
In all experiments, (1) for the classification target tasks, the standard ResNet-50 backbone~\cite{he2016deep} followed by a task-specific classification head is used, (2) for the segmentation target tasks, a U-Net network with a ResNet-50 encoder is used, where the encoder is initialized with the pre-trained models, (3) all target model parameters are fine-tuned, (4) AUC (area under the ROC curve) and Dice coefficient are used for evaluating classification and segmentation target tasks, respectively, (5) mean and standard deviation of performance metrics over ten runs are reported, and (6) statistical analyses based on independent two-sample \emph{t}-test are presented. More implementation details are in Appendix~\ref{sec:appendix_implementation} and project's GitHub page.

\medskip
\noindent\textbf{Pre-trained models:} We benchmark transfer learning from two large-scale natural datasets, ImageNet and iNat2021, and two in-domain medical datasets, CheXpert~\cite{irvin2019chexpert} and ChestX-Ray14~\cite{wang2017chestx}. We pre-train supervised in-domain models which are either initialized randomly or fine-tuned from the ImageNet model. For all other supervised and self-supervised methods, we use existing official and ready-to-use pre-trained models, ensuring that their configurations have been meticulously assembled to achieve the best results in target tasks.

\section{Transfer Learning Benchmarking and Analysis}
\noindent\textbf{1) Pre-trained models on fine-grained data are better suited for segmentation tasks, while pre-trained models on coarse-grained data prevail on classification tasks.}
Medical imaging literature mostly has focused on the pre-training with \emph{coarse-grained} natural image datasets, such as ImageNet~\cite{mustafa2021supervised,WEN2021Rethinking,tajbakhsh2016convolutional,raghu2019transfusion}.
In contrast to previous works, we aim to study the  capability of pre-training with \emph{fine-grained} datasets for transfer learning to medical tasks. 
In fine-grained datasets, visual differences between subordinate classes are often subtle and deeply embedded within local discriminative parts. Therefore, a model has to capture visual details in the local regions for solving a fine-grained recognition task~\cite{Chang2020Devil,Zhuang2020Learning,ZHAO2020Attribute}. We hypothesize that a pre-trained model on a fine-grained dataset derives distinctive local representations that are useful for medical tasks which usually rely upon small, local variations in texture to detect/segment pathologies of interest. To put this hypothesis to the test, we empirically validate how well pre-trained models on large-scale fine-grained datasets can transfer to a range of target medical applications.  This study represents the first effort to rigorously evaluate the impact of pre-training data \textit{granularity} on transfer learning to medical imaging tasks.

\begin{figure*}[t]
\begin{center}
 \includegraphics[width=0.8\textwidth]{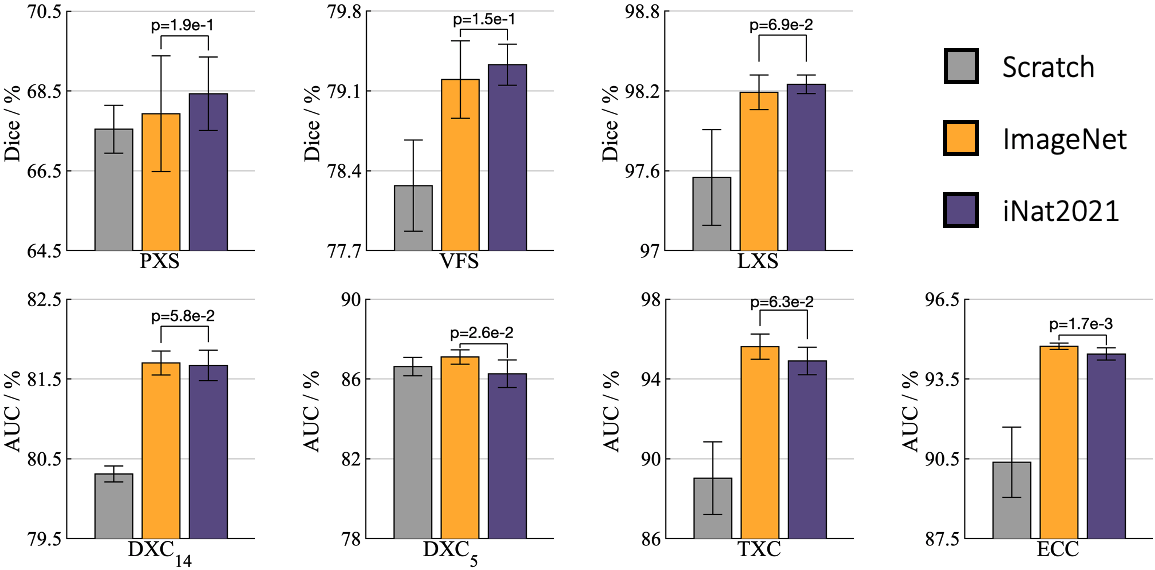}
\end{center}
\caption{
For segmentation (target) tasks (\ie \texttt{PXS}, \texttt{VFS}, and \texttt{LXS}), fine-tuning the model pre-trained on iNat2021 outperforms that on ImageNet, while the model pre-trained on ImageNet prevails on classification (target) tasks (\ie \texttt{DXC$_{14}$}, \texttt{DXC$_5$}, \texttt{TXC}, and \texttt{ECC}), demonstrating the effect of data granularity on transfer learning capability: pre-trained models on the fine-grained data capture subtle features that empowers segmentation target tasks, and pre-trained models on the coarse-grained data encode high-level features that facilitate classification target tasks.
}
\label{fig:imagenet_vs_inat}
\end{figure*}

\smallskip
\noindent\textbf{Experimental setup:} We examine the applicability of iNat2021 as a pre-training source for medical imaging tasks. Our goal is to compare the generalization of the learned features from fine-grained pre-training on iNat2021 with the conventional pre-training on the ImageNet. Given this goal, we use existing official and ready-to-use pre-trained models on these two datasets, and fine-tune them for 7 diverse target tasks, encompassing multi-label classification, binary classification, and pixel-wise segmentation (see \tablename~\ref{tab:datasets}). To provide a comprehensive evaluation, we also include results for training target models from scratch. 

\smallskip
\noindent\textbf{Observations and Analysis:}
As evidenced in \figurename~\ref{fig:imagenet_vs_inat}, fine-tuning from the iNat2021 pre-trained model outperforms the ImageNet counterpart in semantic segmentation tasks, \ie \texttt{PXS}, \texttt{VFS}, and \texttt{LXS}. This implies that, owing to the finer data granularity of iNat2021, the pre-trained model on this dataset yields a  more fine-grained visual feature space, which captures essential pixel-level cues for medical segmentation tasks. This observation gives rise to a natural question of whether this improved performance can be attributed to the larger pre-training data of iNat2021 (2.7M images) compared to ImageNet (1.3M images). In answering this question, we conducted an ablation study on the iNat2021 mini dataset~\cite{vanhorn2021benchmarking} with 500K images to further investigate the impact of data granularity on the learned representations. Our result demonstrates that even with fewer pre-training data, iNat2021 mini pre-trained models can outperform ImageNet counterparts in segmentation tasks (see Appendix~\ref{appendix:mini_inat}). This demonstrates that recovering discriminative features from iNat2021 dataset should be attributed to fine-grained data rather than the larger training data size.

Despite  the  success  of  iNat2021   models  in segmentation  tasks,  fine-tuning  of  ImageNet  pre-trained  features  outperforms iNat2021 in classification tasks, namely \texttt{DXC$_{14}$}, \texttt{DXC$_5$}, \texttt{TXC}, and \texttt{ECC} (see \figurename~\ref{fig:imagenet_vs_inat}). Contrary to our intuition (see \sectionname~\ref{sec:intro}), pre-training on a coarser granularity dataset, such as ImageNet, yields high-level semantic features that are more beneficial for classification tasks.  

\smallskip
\noindent\textbf{Summary:} Fine-grained pre-trained models could be a viable alternative for transfer learning to fine-grained medical tasks, hoping practitioners will find this observation useful in migrating from standard ImageNet checkpoints to reap the benefits we've demonstrated. Regardless of – or perhaps in addition to – other advancements, visually diverse datasets like ImageNet can continue to play a valuable role in building performant medical imaging models.

\medskip
\noindent\textbf{2) Self-supervised ImageNet models outperform supervised ImageNet models.}
A recent family of self-supervised ImageNet models has demonstrated superior transferability in an increasing number of computer vision tasks compared to supervised ImageNet models~\cite{ericsson2021selfsupervised,zhao2021makes,islam2021broad}. Self-supervised models, in particular, capture task-agnostic features that can be easily adapted to different domains~\cite{wei2020semantic,islam2021broad}, while high-level features of supervised pre-trained models may be extraneous when the source and target data distributions are far apart~\cite{zhao2021makes}. We hypothesize this phenomenon is more pronounced in the medical domain, where there is a remarkable domain shift~\cite{ericsson2021selfsupervised} compared to ImageNet. To test this hypothesis, we dissect the effectiveness of a wide range of recent self-supervised methods, encompassing contrastive learning, clustering, and redundancy-reduction methods, on the broadest benchmark yet of various modalities spanning X-ray, CT, and fundus images. This work represents the first effort to rigorously benchmark SSL techniques to a broader range of medical imaging problems.

\begin{figure*}[t]
\begin{center}
 \includegraphics[width=1\textwidth]{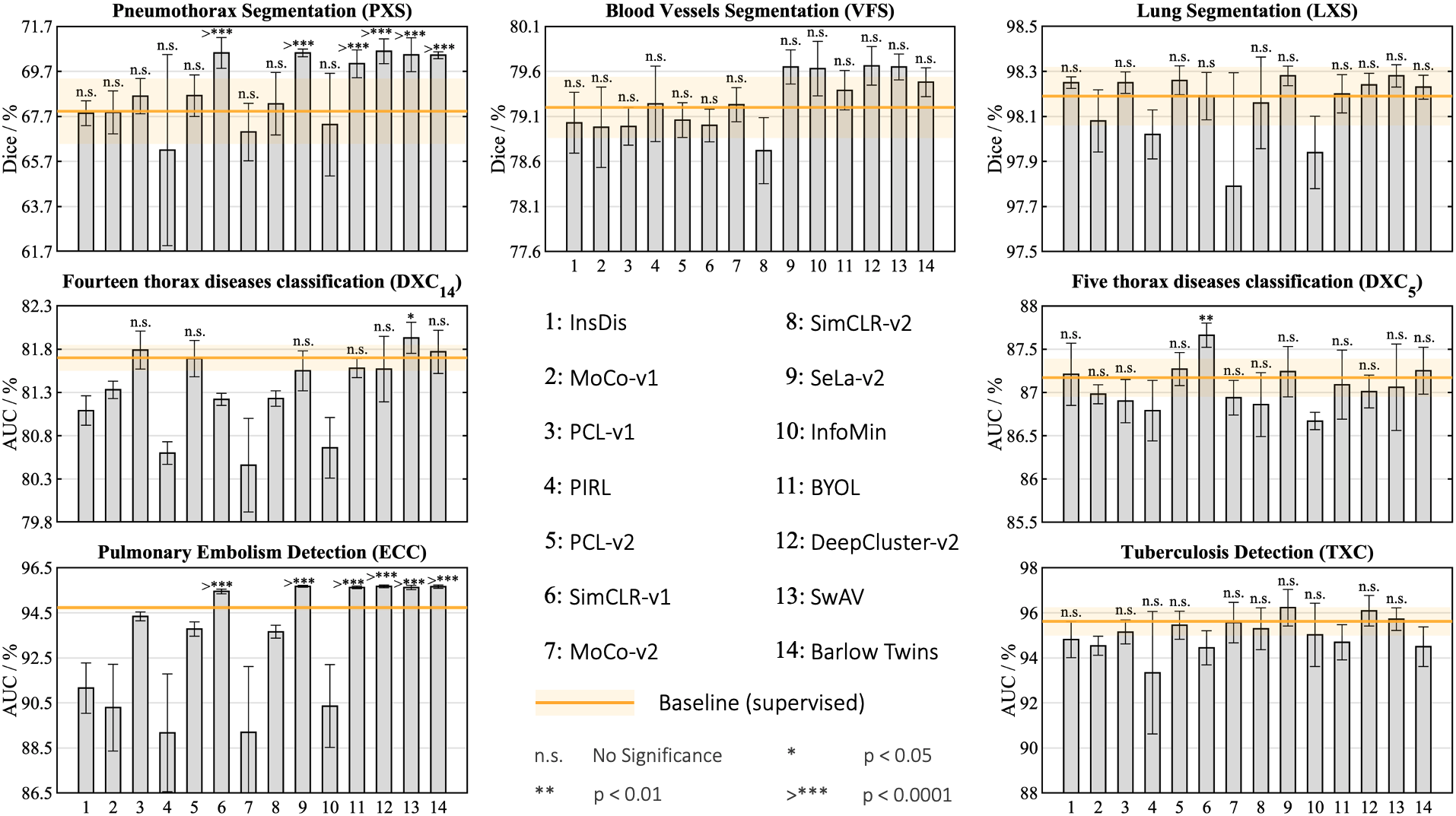}
\end{center}
\caption{
For each target task, in terms of the mean performance, the supervised ImageNet model can be outperformed by at least three self-supervised ImageNet models, demonstrating the higher transferability of self-supervised representation learning. Recent approaches, SwAV~\cite{caron2021unsupervised}, Barlow Twins~\cite{zbontar2021barlow}, SeLa-v2~\cite{caron2021unsupervised}, and DeepCluster-v2~\cite{caron2021unsupervised}, stand out as consistently outperforming the supervised ImageNet model in most target tasks.
We conduct statistical analysis between the supervised model and each self-supervised model in each target task, and show the results for the methods that significantly outperform the baseline or provide comparable performance. Methods are listed in numerical order from left to right. 
}
\label{fig:imagenet_vs_ssl}
\end{figure*}

\smallskip
\noindent\textbf{Experimental setup:}
We evaluate the transferability of 14 popular SSL methods with officially released models, which have been expertly optimized, including contrastive learning (CL) based on instance discrimination (\ie InsDis~\cite{Wu2018insdis}, MoCo-v1~\cite{He2020MocoV1}, MoCo-v2~\cite{chen2020improved}, SimCLR-v1~\cite{Chen2020Simple}, SimCLR-v2~\cite{chen2020big}, and BYOL~\cite{grill2020bootstrap}), CL based on JigSaw shuffling (PIRL~\cite{Misra2020Self}), clustering (DeepCluster-v2~\cite{caron2021unsupervised} and SeLa-v2~\cite{caron2021unsupervised}), clustering bridging CL (PCL-v1~\cite{li2021prototypical}, PCL-v2~\cite{li2021prototypical}, and SwAV~\cite{caron2021unsupervised}), mutual information reduction (InfoMin~\cite{tian2020makes}), and redundancy reduction (Barlow Twins~\cite{zbontar2021barlow}), on 7 diverse medical tasks. All methods are pre-trained on the  ImageNet and use ResNet-50 architecture.
Details of SSL methods can be found in Appendix~\ref{sec:appendix_ssl}. As the baseline, we consider the standard supervised pre-trained model on ImageNet with a ResNet-50 backbone.

\smallskip
\noindent\textbf{Observations and Analysis:}
According to \figurename~\ref{fig:imagenet_vs_ssl}, for each target task, there are at least three self-supervised ImageNet models that outperform the supervised ImageNet model on average. Moreover, the top self-supervised ImageNet models remarkably accelerate the training
process of target models in comparison with supervised counterpart (see Appendix~\ref{sec:appendix_convergence_results}).
Intuitively, supervised pre-training labels encourage the model to retain more domain-specific high-level information, causing the learned representation to be biased toward the pre-training task/dataset's idiosyncrasies. Self-supervised learners, however, capture low/mid level features that are not attuned to domain-relevant semantics, generalizing better to diverse sorts of  target tasks with low-data regimes. 

Comparing the classification (\texttt{DXC$_{14}$}, \texttt{DXC$_5$}, \texttt{TXC}, and \texttt{ECC}) and segmentation tasks (\texttt{PXS}, \texttt{VFS}, and \texttt{LXS}) in \figurename~\ref{fig:imagenet_vs_ssl},  in the latter, a larger number of SSL methods results in better transfer performance, while supervised pre-training falls short. This suggests that when there are larger domain shifts, self-supervised models can provide more precise localization than supervised models.  This is because supervised pre-trained models primarily focus on the smaller discriminative regions of the images, whereas SSL methods attune to larger regions~\cite{ericsson2021selfsupervised,zhao2021makes}, which empowers them with  deriving richer visual information from the entire image. 

\smallskip
\noindent\textbf{Summary:} SSL can learn holistic features more effectively than supervised pre-training, resulting in higher transferability to a variety of medical tasks. It's worth noting that no single SSL method dominates in all tasks, implying that universal pre-training remains a mystery. We hope that the results of this benchmarking, resonating with recent studies in the natural image domain~\cite{ericsson2021selfsupervised,zhao2021makes,islam2021broad}, will lead to more effective transfer learning for medical image analysis.

\medskip
\noindent\textbf{3) Domain-adaptive pre-training bridges the gap between the natural and medical imaging domains.} 
Pre-trained ImageNet models are the predominant standard for transfer learning as they are free, open-source models which can be used for a variety of tasks~\cite{mustafa2021supervised,azizi2021big,haghighi2020learning,WEN2021Rethinking}. Despite the prevailing use of ImageNet models, the remarkable covariate shift between natural and medical images restrain transfer learning~\cite{raghu2019transfusion}. This constraint motivates us to present a practical approach that tailors ImageNet models to medical applications. Towards this end, we investigate domain-adaptive pre-training on natural and medical datasets to tune ImageNet models for medical tasks.

\begin{table*}[t]
\begin{center}
\begin{threeparttable}
\footnotesize
\caption{
Domain-adapted pre-trained models outperform the corresponding ImageNet and in-domain models. For every target task, we performed the independent two sample $t$-test between the best (bolded) vs. others.  Highlighted boxes in green indicate results which have no statistically significant difference at the $p=0.05$ level. When pre-training and target tasks are the same, transfer learning is not applicable, denoted by ``-''. The footnotes compare our results with the state-of-the-art performance for each task.
}
\label{tab:seq_pretraining}
\scriptsize
\begin{tabular*}{\textwidth}
{P{0.26\linewidth} P{0.00001\linewidth} | P{0.00001\linewidth} P{0.135\linewidth} 
P{0.135\linewidth}P{0.135\linewidth}P{0.135\linewidth}P{0.135\linewidth}}
\hline
\multirow{2}{*}{Initialization} &&&\multicolumn{5}{c}{Target tasks}\\
\cline{4-8} \centering & & &
\texttt{DXC$_{14}$}$^a$ & \texttt{DXC$_{5}$}$^b$ & \texttt{TXC}$^{c}$ &\texttt{PXS}$^d$&\texttt{LXS}$^{e}$\\
\hline
Scratch &&& 80.31$\pm$0.10 &	86.60$\pm$0.17&	89.03$\pm$1.82&	67.54$\pm$0.60&	97.55$\pm$0.36 \\ 
\hline
ImageNet&&& 81.70$\pm$0.15&	\cellcolor{maroon!15} 87.10$\pm$0.36&	95.62$\pm$0.63&	67.93$\pm$1.45&	\cellcolor{maroon!15} 98.19$\pm$0.13\\
\hline
ChestX-ray14~\cite{wang2017chestx} &&& - & \cellcolor{maroon!15} \textbf{87.40$\pm$0.26}&	96.32$\pm$0.65&	\cellcolor{maroon!15} 68.92$\pm$0.98&	98.18$\pm$0.06\\
CheXpert~\cite{irvin2019chexpert} &&& 81.99$\pm$0.08& - &97.07$\pm$0.95&	\cellcolor{maroon!15} 69.30$\pm$ 0.50& \cellcolor{maroon!15}	98.25$\pm$0.04 \\
\hline
ImageNet$\rightarrow$ChestX-ray14 &&& - & \cellcolor{maroon!15} 87.09$\pm$0.44&	\cellcolor{maroon!15}\textbf{98.47$\pm$0.26}&	\cellcolor{maroon!15}\textbf{69.52$\pm$0.38}&	\cellcolor{maroon!15} 98.27$\pm$0.03\\
ImageNet$\rightarrow$CheXpert&&& \cellcolor{maroon!15}\textbf{82.25$\pm$0.18}& - &97.33$\pm$0.26& \cellcolor{maroon!15}	69.36$\pm$0.49&	\cellcolor{maroon!15}\textbf{98.31$\pm$0.05}\\

\hline
\end{tabular*}
\begin{tablenotes}
 \scriptsize
        \item $^a$ ~\cite{Kim2021XProtoNet} holds an AUC of $82.00\%$ vs. 82.25\%$\pm$0.18\% (ours) 
        \item $^b$ ~\cite{PHAM2021Interpreting} holds an AUC of $89.40\%$ w/ disease
dependencies (DD) vs. 87.40\%$\pm$0.26\% (ours w/o DD)
        \item $^{c}$ ~\cite{Rajaraman2021Chest} holds an AUC of $95.35\%\pm1.86\%$ vs. \textbf{$98.47\%\pm0.26\%$} (ours)
        \item $^{d}$ ~\cite{haghighi2021transferable} holds a Dice of $68.41\%\pm0.14\%$ vs. \textbf{$69.52\%\pm0.38\%$} (ours)
        \item $^{e}$ ~\cite{Reamaroon2021Robust} holds a Dice of  $96.94\%\pm2.67\%$ vs. \textbf{$98.31\%\pm0.05\%$} (ours)
    \end{tablenotes}
    \end{threeparttable}
    \end{center}
\end{table*}

\smallskip
\noindent\textbf{Experimental Setup:} 
The domain-adaptive paradigm originated from natural language processing~\cite{gururangan2020dont}. This is a sequential pre-training approach in which a model is first pre-trained on a massive general dataset, such as ImageNet, and then pre-trained on domain-specific datasets, resulting in domain-adapted pre-trained models. For the first pre-training step, we used the supervised ImageNet model. For the second pre-training step, we created two new models that were initialized through the ImageNet model followed by supervised pre-training on CheXpert (ImageNet$\rightarrow$CheXpert) and ChestX-ray14 (ImageNet$\rightarrow$ChestX-ray14). We compare the domain-adapted models with (1) the ImageNet model, and (2) two supervised pre-trained models on CheXpert and ChestX-ray14, which are randomly initialized. In contrast to previous work~\cite{azizi2021big} which is limited to two classification tasks, we evaluate domain-adapted models on a broader range of five target tasks on chest X-ray scans; these tasks span classification and segmentation, ascertaining the generality of our findings.

\smallskip
\noindent\textbf{Observations and Analysis:} We draw the following observations from \tablename~\ref{tab:seq_pretraining}. (1) Both ChestX-ray14 and CheXpert models consistently outperform the ImageNet model in all cases. This observation implies that in-domain medical transfer learning, whenever possible, is  preferred over ImageNet transfer learning. Our conclusion is opposite to~\cite{WEN2021Rethinking}, where in-domain pre-trained models outperform ImageNet models in controlled setups but lag far behind the real-world ImageNet models.  
(2) The overall trend showcases the advantage of domain-adaptive pre-training. Specifically, for \texttt{DXC$_{14}$}, fine-tuning the ImageNet$\rightarrow$CheXpert model surpasses both ImageNet and CheXpert models. Furthermore, the dominance of domain-adapted models (ImageNet$\rightarrow$CheXpert and ImageNet$\rightarrow$ChestX-ray14) over ImageNet and corresponding in-domain models (CheXpert and ChestX-ray14) is conserved at \texttt{LXS}, \texttt{TXC}, and \texttt{PXS}.
This suggests that domain-adapted models leverage the learning experience of the ImageNet model and further refine it with domain-relevant data,  resulting in more pronounced representation. 
(3)  In \texttt{DXC$_{5}$}, the domain-adapted performance decreases relative to corresponding ImageNet and in-domain models. This is most likely due to the lesser number of images in the in-domain pre-training dataset than the target dataset (75K vs. 200K), suggesting that  in-domain pre-training data should be larger than the target data~\cite{gururangan2020dont,reed2021selfsupervised}. 

\smallskip
\noindent\textbf{Summary:} Continual pre-training can bridge the domain gap between natural and medical images. Concretely, we leverage the readily conducted annotation efforts to produce more performant medical imaging models and reduce future annotation burdens. We hope our findings posit new research directions for developing specialized pre-trained models in medical imaging. Our pre-trained models, in-domain (CheXpert and ChestX-ray14) as well as domain-adapted (ImageNet$\rightarrow$CheXpert and ImageNet$\rightarrow$ChestX-ray14), are publicly available on our GitHub page.

\section{Conclusion and Future Work}
We provide the first fine-grained and up-to-date study on the transferability of
various brand-new pre-training techniques for medical imaging tasks, answering central and timely questions on transfer learning in medical image analysis. Our empirical evaluation suggests that: (1) what truly matters for the segmentation tasks is fine-grained representation rather than high-level semantic features, (2) top self-supervised ImageNet models outperform the supervised ImageNet model, offering a new transfer learning standard for medical imaging, and (3) ImageNet models can be strengthened with continual in-domain pre-training.

\noindent\textbf{Future work:} In this work, we have considered transfer learning from the supervised ImageNet model as the baseline, on which all our evaluations are benchmarked. To compute p-values for statistical analysis, 14 SSL, 5 supervised, and 2 domain-adaptive pre-trained models were run 10 times each on a set of 7 target tasks--- leading to a large number of experiments (1,420).
Nevertheless, our self-supervised models were all pre-trained on ImageNet with ResNet50 as the backbone. While ImageNet is generally regarded as a strong source for pre-training~\cite{vanhorn2021benchmarking,WEN2021Rethinking}, pre-training modern self-supervised models with iNat2021 and in-domain medical image data on various architectures may offer even deeper insights into transfer learning for medical imaging.

\subsubsection{Acknowledgments}
This research has been supported partially by ASU and Mayo Clinic through a Seed Grant and an Innovation Grant, and partially by the NIH under Award Number R01HL128785.  The content is solely the responsibility of the authors and does not necessarily represent the official views of the NIH. This work has utilized the GPUs provided partially by the ASU Research Computing and partially by the Extreme Science and Engineering Discovery Environment (XSEDE) funded by the National Science Foundation (NSF) under grant number ACI-1548562. We thank Nahid Islam for evaluating the self-supervised methods on the PE detection target task.  The content of this paper is covered by patents pending. 

%
%
%
\bibliographystyle{splncs04}

\newpage

\title{Supplementary Material}

\appendix

\section*{Appendix} 

\section{Datasets}
\label{sec:appendix_datasets}
\noindent\textbf{iNat2021~\cite{vanhorn2021benchmarking}:} The  iNaturalist2021 dataset (iNat2021) is a recent large-scale, fine-grained species dataset with 2.7M training images covering 10k species. This dataset facilitates fine-grained visual classification problems. Compared to the more widely used dataset, ImageNet, iNat2021 contains a greater number of these fine-grained images but a narrower range of visual diversity.

\noindent\textbf{iNat2021 mini~\cite{vanhorn2021benchmarking}:}
In addition to the full sized dataset, Horn \etal~\cite{vanhorn2021benchmarking} created
a smaller version of iNat2021, named iNat2021 mini, that contains 50 training images per species, sampled from the full train split. In total, iNat2021 mini includes 500K training images covering 10k species.

\noindent\textbf{ChestX-ray14~\cite{wang2017chestx}:} This hospital-scale chest X-ray dataset contains 112K frontal-view X-ray images taken from a sample of 30K unique patients. ChestX-ray14 provides an official patient-wise split for training (86K images) and test sets (25K images). In this dataset, 51K images have at least one of the 14 thorax diseases. We use the official data split and report the mean AUC score over 14 diseases for the multi-label chest X-ray classification task.

\noindent\textbf{CheXpert~\cite{irvin2019chexpert}:} This large-scale publicly available dataset contains 224K high-quality chest X-ray images taken from a sample of 65K patients. 
The training images were annotated by a labeler to automatically
detect the presence of 14 thorax diseases in radiology reports, capturing uncertainties inherent
in radiograph interpretation. The test set consists of 234 images from 200 patients.  The test images were manually annotated by board-certified radiologists for 5 selected diseases, i.e., Cardiomegaly, Edema, Consolidation, Atelectasis, and Pleural Effusion. We use the official data split and report the mean AUC score over 5 test diseases.

\noindent\textbf{SIIM-ACR Pneumothorax Segmentation~\cite{PNEchallenge}:} The Society for Imaging Informatics in Medicine (SIIM) and American College of Radiology provided the SIIM-ACR Pneumothorax Segmentation dataset, consisting of 10K chest X-ray images and the segmentation masks for Pneumothorax disease. We randomly divided the dataset into training (80\%) and testing (20\%), and the segmentation performance was evaluated by using the Dice coefficient score.

\noindent\textbf{RSNA PE Detection~\cite{PEchallenge}:} This dataset is the largest publicly available annotated Pulmonary Embolism (PE) dataset, comprised of more than 7K CT scans with a varying number of images in each scan. Each image has  been  annotated  for  the  presence  or  absence  of  the  PE. Also, each scan has been labeled for additional nine patient-level labels. We randomly split the data at patient-level to training (6K) and testing (1K) sets, respectively. Correspondingly,  there  are 1.5M and 248K images in  the training  and testing sets, respectively. We report the AUC score for the PE detection task.

\noindent\textbf{NIH Shenzhen CXR~\cite{Jaeger2014Tow}:} The dataset contains 662 frontal-view chest X-rays, of which 326 are normal cases and 336 are cases with manifestations of Tuberculosis (TB), including pediatric X-rays (AP). We randomly divide the dataset into a training set (80\%) and a test set (20\%). We report the AUC score for the Tuberculosis detection task.

\noindent\textbf{NIH Montgomery~\cite{Jaeger2014Tow}:} The dataset contains 138 frontal-view chest X-rays from Montgomery County’s Tuberculosis screening program, of which 80 are normal cases and 58 are cases with manifestations of TB. The segmentation masks for left and right lungs are provided.  We randomly divided the dataset into a training set (80\%) and a test set (20\%) and report the mean Dice score for the lung segmentation task.

\noindent\textbf{DRIVE~\cite{Budai2013Robust}:} The dataset contains 40 retinal images, separated by its providers into a training set (20 images) and a test set (20 images). For all images, manual segmentation of the vasculature is provided. We use the official data split and report the mean Dice score for the segmentation of blood vessels.

\section{Implementation}
\label{sec:appendix_implementation}
We evaluate popular publicly available representations that have been pre-trained with various methods and datasets across a variety of target tasks. Therefore, we control other influencing factors such as pre-processing, network architecture, and transfer hyperparameters. We run each method ten times on all of the target tasks and
report the average, standard deviation, and further present statistical analysis based on an independent two-sample \textit{t}-test.

 \noindent\textbf{Architecture:} We fix the network architecture in all experiments since we seek to understand the competitiveness of representations rather than benefits from architecture. Therefore, all the pre-trained models leverage the same ResNet-50 backbone. For transfer learning to the classification target tasks, we take the pre-trained ResNet-50 models and append a task-specific classification head. For the segmentation target tasks, we utilize a U-Net\footnote{\label{foot:densenet121}Segmentation Models: \href{https://github.com/qubvel/segmentation_models.pytorch}{https://github.com/qubvel/segmentation\_models.pytorch}} network with a ResNet-50 encoder, where the encoder is initialized with the pre-trained models. We have evaluated the transfer learning performance of   all pre-trained models by fine-tuning all layers in the downstream networks.

\noindent\textbf{Preprocessing and data augmentation:} For target tasks on X-ray modality (\texttt{DXC$_{14}$}, \texttt{DXC$_5$}, \texttt{TXC}, \texttt{LXS}, and \texttt{PXS}), Fundoscopic modality  (\texttt{VFS}), and CT modality (\texttt{PCC}), we resize the images to 224$\times$224, 512$\times$512, and 576$\times$576, respectively. For all classification target tasks, we apply standard data augmentation techniques, including random cropping, horizontal flipping, and rotating. For segmentation tasks on X-ray modality (\texttt{LXS} and \texttt{PXS}), we employ RandomBrightnessContrast, RandomGamma, OpticalDistortion, elastic transformation, and grid distortion. For segmentation task on fundoscopic modality (\texttt{VFS}), we use random rotation, Gaussian noise, color jittering, and horizontal, vertical and diagonal flips.

\noindent\textbf{Training parameters:}
Since different datasets require different optimal settings, we strive to optimize each target task with the best performing hyper-parameters. 
In all experiments, we use Adam optimizer with $\beta_1= 0.9$, $\beta_2= 0.999$. We use \emph{ReduceLROnPlateau} and \emph{cosine} learning rate decay schedulers for classification and segmentation tasks, respectively; if no improvement is seen in the validation set for a certain number of epochs, the learning rate is reduced. We employ early-stop mechanism using  the 10\% of the training data as the validation set to avoid over-fitting.
For X-ray classification tasks (\texttt{DXC$_{14}$}, \texttt{DXC$_5$}, and \texttt{TXC}), segmentation tasks (\texttt{VFS}, \texttt{LXS}, and \texttt{PXS}), and PE detection task (\texttt{ECC}),  we use a learning rate of $2e-4$, $1e-3$, and $4e-4$, respectively.

\begin{table*}[t]
\begin{center}
\begin{threeparttable}
\footnotesize
\caption{
Evaluation of iNat2021 mini dataset on segmentation medical tasks. Even with less than half number of pre-training samples, iNat2021 mini achieves equal or superior performance over ImageNet counterpart. Best performance is \textbf{bolded} and second best is \underline{underlined}.
}
\label{tab:appendix_abl_mini_inat}
\begin{tabular}
{P{0.17\linewidth} P{0.2\linewidth}  P{0.00001\linewidth} | P{0.00001\linewidth} P{0.17\linewidth} 
P{0.17\linewidth} P{0.17\linewidth} }
\hline
\multicolumn{2}{c}{Pre-training task} &&&\multicolumn{3}{c}{Target tasks}\\
\cline{1-2} \cline{5-7} \centering Dataset& \#training data & & &
\texttt{PXS} & \texttt{VFS} & \texttt{LXS} \\
\hline
ImageNet&1.3M &&&67.93$\pm$1.45 &79.20$\pm$0.34&\underline{98.19$\pm$0.13} \\
iNat2021 mini&500K &&& \underline{68.26$\pm$1.48}&\underline{79.24$\pm$0.28}&\underline{98.19$\pm$0.09} \\
iNat2021&2.7M &&& \textbf{68.43$\pm$0.92} &\textbf{79.33$\pm$0.18}& \textbf{98.25$\pm$0.07}\\
\hline
\end{tabular}
    \end{threeparttable}
    \end{center}
\end{table*}

\section{Ablation study on iNat2021 mini dataset}
\label{appendix:mini_inat}
We further investigate the capability of  pre-trained models on fine-grained datasets in capturing fine-grained details by examining iNat2021 mini dataset for segmentation tasks. iNat2021 mini contains 500K images, which is less than half compared to ImageNet. The results in \tablename~\ref{tab:appendix_abl_mini_inat} indicate that even with fewer training data, iNat2021 achieves equal or better performance than ImageNet counterpart. This observation suggests that the superior performance of iNat2021 over ImageNet pre-trained model in segmentation tasks should be attributed to the fine-grained nature of data rather than larger pre-training size. 

\begin{table*}[t]
\begin{center}
\begin{threeparttable}
\footnotesize
\caption{
Benchmarking transfer learning from supervised iNat2021 and ImageNet models on seven medical tasks. Pre-trained models on iNat2021 are better suited for segmentation tasks (\ie \texttt{LXS}, \texttt{VFS}, and \texttt{PXS}), while pre-trained models on ImageNet prevail on classification tasks (\ie \texttt{DXC$_{14}$}, \texttt{DXC$_5$}, \texttt{TXC}, and \texttt{ECC}). The best model in each application is \textbf{bolded}.
}
\label{tab:inat_vs_imagenet}
\begin{tabular*}{\textwidth}
{p{0.48\linewidth} P{0.00001\linewidth} | P{0.00001\linewidth} P{0.15\linewidth}  P{0.15\linewidth}P{0.15\linewidth}}
\hline
\centering \multirow{2}{*}{ Downstream task}&&&\multicolumn{3}{c}{Initialization}\\
 & & & Random & ImageNet & iNat2021\\
\hline
Pneumothorax segmentation (\texttt{PXS})&&& 67.54$\pm$0.60&67.93$\pm$1.45 & \textbf{68.43$\pm$0.92} \\
Lung segmentation (\texttt{LXS})&&& 97.55$\pm$0.36 & 98.19$\pm$0.13 & \textbf{98.25$\pm$0.07}\\
Blood Vessels Segmentation (\texttt{VFS}) &&& 78.27$\pm$0.40&
79.20$\pm$0.34&\textbf{79.33$\pm$0.18} \\
14 thorax diseases classification (\texttt{DXC$_{14}$})&&& 80.31$\pm$0.10 & \textbf{81.70$\pm$0.15} & 81.67$\pm$0.19\\
5 thorax disease classification (\texttt{DXC$_{5}$})&&& 86.62$\pm$0.46& \textbf{87.10$\pm$0.36}& 86.26$\pm$0.69 \\
Tuberculosis Detection (\texttt{TXC})&&&  89.03$\pm$1.82& \textbf{95.62$\pm$0.63} & 94.90$\pm$0.69 \\
Pulmonary Embolism Detection (\texttt{ECC})&&& 90.37$\pm$1.32 & \textbf{94.73$\pm$0.12} & 94.44$\pm$0.23\\
\hline
\end{tabular*}
\end{threeparttable}
\end{center}
\end{table*}

\section{Tabular results}
\label{sec:appendix_tab_results}
In this section, tabulated results of different experiments are reported. The results of Fig.~\ref{fig:imagenet_vs_inat} and Fig.~\ref{fig:imagenet_vs_ssl} in the main paper are presented in \tablename~\ref{tab:inat_vs_imagenet} and \tablename~\ref{tab:ssl_vs_imagenet}, respectively.

\begin{table*}[t]
\begin{center}
\begin{threeparttable}
\caption{
Benchmarking transfer learning from fourteen self-supervised ImageNet pre-trained models on seven medical tasks. Self-supervised ImageNet models outperform supervised ImageNet models. 
The best model is \textbf{bolded}, and all the other models that outperform supervised baseline are \underline{underlined}.
}
\label{tab:ssl_vs_imagenet}
\scriptsize
\begin{tabular*}{\textwidth}
{p{0.18\linewidth} P{0.00001\linewidth} | P{0.00001\linewidth} P{0.18\linewidth}  P{0.18\linewidth}P{0.18\linewidth}P{0.18\linewidth}}
\hline
\centering \multirow{2}{*}{Pre-training}  &&&\multicolumn{4}{c}{Downstream task}\\
 & & & \texttt{DXC$_{14}$} & \texttt{DXC$_{5}$} & \texttt{TXC} & \texttt{ECC} \\
 \hline
 Supervised &&& 81.70$\pm$0.15 & 87.10$\pm$0.36 & 95.62$\pm$0.63 & 94.73$\pm$0.12 \\
 \hline
InsDis &&& 81.09$\pm$0.17 & \underline{87.21$\pm$0.36}&94.81$\pm$0.73&91.16$\pm$1.12
\\
MoCo-v1 &&& 81.33$\pm$0.10  &86.98$\pm$0.11&94.54$\pm$0.42&90.29$\pm$1.92
	\\
PCL-v1 &&& \underline{81.79$\pm$0.22} &86.90$\pm$0.25&95.15$\pm$0.53&94.34$\pm$0.20
	\\
PIRL &&& 80.60$\pm$0.13&86.79$\pm$0.35&93.34$\pm$2.72&89.17$\pm$2.62
	\\
PCL-v2 &&& 81.69$\pm$0.21&\underline{87.27$\pm$0.19}&95.45$\pm$0.62&93.78$\pm$0.31
	\\
SimCLR-v1 &&& 81.22$\pm$0.07&\textbf{87.66$\pm$0.14}&94.45$\pm$0.76&\underline{95.45$\pm$0.11}
	\\
MoCo-v2 &&&80.46$\pm$0.54&86.94$\pm$0.20&95.57$\pm$0.90&89.20$\pm$2.92
	\\
SimCLR-v2&&&81.23$\pm$0.09&86.86$\pm$0.37&95.29$\pm$0.93&93.66$\pm$0.29
	\\
SeLa-v2&&& 81.55$\pm$0.23&\underline{87.24$\pm$0.29}&\textbf{96.23$\pm$0.81}&\textbf{95.68$\pm$0.05}
	\\
InfoMin&&&80.66$\pm$0.35&86.67$\pm$0.10&95.02$\pm$1.40&90.36$\pm$1.84
	\\
BYOL&&&81.58$\pm$0.11&87.09$\pm$0.40&94.69$\pm$0.78&\underline{95.63$\pm$0.05}
	\\
DeepCluster-v2&&&81.57$\pm$0.38&87.01$\pm$0.19&\underline{96.09$\pm$0.68}&\textbf{95.68$\pm$0.06}
	\\
SwAV&&&\textbf{81.93$\pm$0.18}&87.06$\pm$0.50&\underline{95.72$\pm$0.50}&\underline{95.63$\pm$0.10}
	\\
Barlow Twins&&&\underline{81.77$\pm$0.25}&\underline{87.25$\pm$0.27}&94.50$\pm$0.88&\underline{95.66$\pm$0.07}
	\\

\hline
\hline
 & & & \texttt{PXS} & \texttt{LXS} &\texttt{VFS} \\
 \hline
  Supervised &&& 67.93$\pm$1.45 & 98.19$\pm$0.13 & 79.20$\pm$0.34 \\
  \hline
  InsDis &&&67.84$\pm$0.55&\underline{98.25$\pm$0.03}&79.03$\pm$0.34
  \\
MoCo-v1 &&&67.88$\pm$0.95&98.08$\pm$0.14&78.98$\pm$0.45
\\
PCL-v1 &&&\underline{68.60$\pm$0.78}&\underline{98.25$\pm$0.05}&78.99$\pm$0.21
\\
PIRL &&&66.20$\pm$4.24&98.02$\pm$0.11&\underline{79.24$\pm$0.42}
\\
PCL-v2 &&&\underline{68.62$\pm$0.92}&\underline{98.26$\pm$0.06}&79.06$\pm$0.19
\\
SimCLR-v1 &&&\underline{70.52$\pm$0.69}&98.19$\pm$0.10&79.00$\pm$0.18
\\
MoCo-v2 &&&67.01$\pm$1.28&97.79$\pm$0.50&\underline{79.23$\pm$0.19}
\\
SimCLR-v2&&&\underline{68.26$\pm$1.39}&98.16$\pm$0.20&78.72$\pm$0.37
\\
SeLa-v2&&&\underline{70.52$\pm$0.17}&\textbf{98.28$\pm$0.04}&\underline{79.65$\pm$0.19}
\\
InfoMin&&&67.34$\pm$2.28&97.94$\pm$0.16&\underline{79.63$\pm$0.30}
\\
BYOL&&&\underline{70.04$\pm$0.62}&\underline{98.20$\pm$0.08}&\underline{79.39$\pm$0.22}
\\
DeepCluster-v2&&&\textbf{70.59$\pm$0.55}&\underline{98.24$\pm$0.05}&\textbf{79.66$\pm$0.21}
\\
SwAV&&&\underline{70.44$\pm$0.75}&\textbf{98.28$\pm$0.05}&\underline{79.65$\pm$0.14}
\\
Barlow Twins&&&\underline{70.42$\pm$0.15}&\underline{98.23$\pm$0.05}&\underline{79.48$\pm$0.16}
\\
 
\hline
\end{tabular*}
\end{threeparttable}
\end{center}
\end{table*}

\begin{table*}[t]
\begin{center}
\begin{threeparttable}
\scriptsize
\caption{
Fine-tuning from iNat2021 model provides higher performance in all segmentation tasks and considerably accelerates the training process in two out of three tasks in comparison to the ImageNet counterpart. The average performance and number of training epochs over ten runs is reported for each model in each target task. The best performance in each task is bolded.
}
\label{tab:imagenet_vs_inat_convergence}
\begin{tabular*}{\textwidth}
{p{0.16\linewidth} P{0.00001\linewidth} | P{0.00001\linewidth} P{0.125\linewidth} P{0.115\linewidth} P{0.00001\linewidth} | P{0.00001\linewidth} P{0.125\linewidth}  P{0.135\linewidth} P{0.00001\linewidth} | P{0.00001\linewidth} P{0.12\linewidth}P{0.115\linewidth}}
\hline
\centering \multirow{2}{*}{Initialization} & & &   \multicolumn{2}{c} {\texttt{PXS}} &&& \multicolumn{2}{c} {\texttt{VFS}}&&&\multicolumn{2}{c}{\texttt{LXS}}\\
 \cline{4-5}\cline{8-9}\cline{12-13} & & & Dice($\uparrow$) & \#Epochs($\downarrow$)&&& Dice($\uparrow$) & \#Epochs($\downarrow$) &&& Dice($\uparrow$) & \#Epochs($\downarrow$)\\
 \hline
 Random&&& 67.54$\pm$0.60&46.0$\pm$13.87&&&78.27$\pm$0.40&100.0$\pm$0.0&&&97.55$\pm$0.36& 92.6$\pm$51.38\\ 
 \hline
 ImageNet & && 67.93$\pm$1.45 &45.9$\pm$28.25&& & 79.20$\pm$0.34&71$\pm$18.29&&&98.19$\pm$0.13& 84.9$\pm$27.55\\
 \hline
 iNat2021&&& \textbf{68.43$\pm$0.92}&41.8$\pm$17.98&&&\textbf{79.33$\pm$0.18}&59.3$\pm$5.58&&& \textbf{98.25$\pm$0.07}&98.9$\pm$26.41\\
\hline
\end{tabular*}
\end{threeparttable}
\end{center}
\end{table*}

\begin{table*}[t]
\begin{center}
\begin{threeparttable}
\scriptsize
\caption{
Fine-tuning from the best self-supervised models provide significantly better or equivalent performance and accelerate the training process in comparison to the supervised counterpart. The average performance and number of training epochs over ten runs is reported for each model in each target task. The best performance in each task is bolded. 
}
\label{tab:ssl_vs_imagenet_convergence}
\begin{tabular*}{\textwidth}
{p{0.16\linewidth} P{0.00001\linewidth} | P{0.00001\linewidth} P{0.125\linewidth} P{0.115\linewidth} P{0.00001\linewidth} | P{0.00001\linewidth} P{0.125\linewidth}  P{0.135\linewidth} P{0.00001\linewidth} | P{0.00001\linewidth} P{0.115\linewidth}P{0.11\linewidth}}
\hline
\centering \multirow{2}{*}{Initialization} & & &  \multicolumn{2}{c}{\texttt{DXC$_{14}$}} &&& \multicolumn{2}{c} {\texttt{PXS}} &&& \multicolumn{2}{c} {\texttt{VFS}} \\
 \cline{4-5}\cline{8-9}\cline{12-13} & & & AUC($\uparrow$) & \#Epochs($\downarrow$)&&& Dice($\uparrow$) & \#Epochs($\downarrow$) &&& Dice($\uparrow$) & \#Epochs($\downarrow$)\\
 \hline
 Random&&& 80.40$\pm$0.05& 68.2$\pm$5.07&&&67.54$\pm$0.60&46.0$\pm$13.87&&&78.27$\pm$0.40&100.0$\pm$0.0\\ 
 \hline
 Supervised & && 81.70$\pm$0.15 & 34.2$\pm$4.32&&&67.93$\pm$1.45 &45.9$\pm$28.25&& & 79.20$\pm$0.34&71$\pm$18.29
	\\
 \hline
 SeLa-v2&&&81.55$\pm$0.23&10.0$\pm$0.71&&& 70.52$\pm$0.17&39.40$\pm$8.99&&& 79.65$\pm$0.19& 45$\pm$10.27
	\\
DeepCluster-v2&&&81.57$\pm$0.38&  8.80$\pm$1.92&&& \textbf{70.59$\pm$0.55}& 37.0$\pm$17.46&&&  \textbf{79.66$\pm$0.21}& 43.6$\pm$3.58\\
SwAV&&& \textbf{81.93$\pm$0.18}&13.0$\pm$2.55&&& 70.44$\pm$0.75&44.8$\pm$18.31&&& 79.65$\pm$0.14&44.4$\pm$5.27\\
Barlow Twins&&& 81.77$\pm$0.25&12.4$\pm$2.61&&& 70.42$\pm$0.15&55.8$\pm$24.32&&& 79.48$\pm$0.16&47.6$\pm$4.72\\
\hline
\end{tabular*}
\end{threeparttable}
\end{center}
\end{table*}

\begin{table*}[t]
\begin{center}
\begin{threeparttable}
\scriptsize
\caption{
Fine-tuning from the domain-adapted pre-trained models provides higher performance in all tasks and speeds up the training process compared to the corresponding ImageNet models in most cases. The average performance and number of training epochs over ten runs is reported for each model in each target task. The best performance in each task is bolded. ``CXR14'' denotes the ChestX-ray14 dataset. When pre-training and target tasks are the same, transfer learning is not applicable, denoted by ``-''.
}
\label{tab:domain_adapted_vs_imagenet_convergence}
\begin{tabular*}{\textwidth}
{p{0.21\linewidth} P{0.00001\linewidth} | P{0.00001\linewidth} P{0.115\linewidth} P{0.11\linewidth} P{0.000001\linewidth} | P{0.000001\linewidth} P{0.12\linewidth}  P{0.11\linewidth} P{0.00001\linewidth} | P{0.000001\linewidth} P{0.11\linewidth}P{0.13\linewidth}}
\hline
\centering \multirow{2}{*}{Initialization} & & &  \multicolumn{2}{c}{\texttt{DXC$_{14}$}} &&& \multicolumn{2}{c} {\texttt{PXS}} &&& \multicolumn{2}{c} {\texttt{LXS}} \\
 \cline{4-5}\cline{8-9}\cline{12-13} & & & AUC($\uparrow$) & \#Epochs($\downarrow$)&&& Dice($\uparrow$) & \#Epochs($\downarrow$) &&& Dice($\uparrow$) & \#Epochs($\downarrow$)\\
 \hline
 Random&&& 80.31$\pm$0.10& 68.2$\pm$5.07&&&67.54$\pm$0.60&46.0$\pm$13.87&&&97.55$\pm$0.36& 92.6$\pm$51.38\\ 
 \hline
 ImageNet & && 81.70$\pm$0.15 & 34.2$\pm$4.32&&&67.93$\pm$1.45 &45.9$\pm$28.25&& & 98.19$\pm$0.13& 84.9$\pm$27.55\\
 \hline
CXR14  &&& - &- &&& 68.92$\pm$0.98&49.0$\pm$35.48&&&98.18$\pm$0.06&66.2$\pm$20.58\\
CheXpert&&& 81.99$\pm$0.08& 15.8$\pm$4.32&&& 69.30$\pm$ 0.50&42.2$\pm$22.85&&&  98.25$\pm$0.04 &84.0$\pm$16.85\\
\hline
ImageNet$\rightarrow$CXR14 &&& -&- &&&  		\textbf{69.52$\pm$0.38}&37.6$\pm$15.49&&&	 98.27$\pm$0.03&72.6$\pm$30.92\\
ImageNet$\rightarrow$CheXpert&&& \textbf{82.25$\pm$0.18}&22.2$\pm$3.49&&& 69.36$\pm$0.49&45.8$\pm$5.93&&&	\textbf{98.31$\pm$0.05}&110.7$\pm$40.08\\
\hline
\end{tabular*}
\end{threeparttable}
\end{center}
\end{table*}

\section{Convergence Time Analysis}
\label{sec:appendix_convergence_results}
Transfer learning  attracts great attention since it improves the target performance and accelerates the model convergence when compared to training from scratch. In that respect, a good pre-trained model should yield better target performance with less training time. Therefore, we further evaluate the pre-trained models in terms of accelerating the training process of various medical tasks. In the following, we provide the training time results for each of the three groups of experiments in the main paper. We used the early-stop technique in all target tasks, and report the average number of training epochs over ten runs for each model. 

\smallskip
\noindent\textbf{1) Supervised ImageNet model vs. supervised iNat2021 model.} We provide the training time of the segmentation tasks in which the iNat2021 model outperforms its ImageNet counterpart. The results in \tablename~\ref{tab:imagenet_vs_inat_convergence} indicate that fine-tuning from the iNat2021 model provides higher performance in all segmentation tasks and considerably accelerates the training process in two out of three tasks in comparison to the ImageNet counterpart. 

\smallskip
\noindent\textbf{2) Supervised ImageNet model vs. self-supervised ImageNet models.}
We compare the training time of the top four self-supervised ImageNet models (based on the overall performances in different target tasks) to the supervised ImageNet model in three target tasks, including classification and segmentation. To provide a comprehensive evaluation, we also include results for training target models from scratch. 

Our results in \tablename~\ref{tab:ssl_vs_imagenet_convergence} demonstrate that fine-tuning from the best self-supervised models in each target task provide significantly better or equivalent performance and remarkably accelerate the training process in comparison to the supervised counterpart. Specifically, in \texttt{DXC$_{14}$} task, SwAV and Barlow Twins achieve superior performance with significantly less number of training epochs compared to supervised ImageNet model. Similarly, in \texttt{PXS} task, SeLa-v2, DeepCluster-v2, and SwAV outperform supervised ImageNet model in terms of both performance and training time. Furthermore, in \texttt{VFS} task, all the self-supervised models yield higher performance with less training time compared to supervised ImageNet model. 

Additionally, considering the principle that a good representation should generalize to multiple target tasks with limited fine-tuning~\cite{goyal2019scaling}, we fine-tuned all the models for the same number of training epochs in \texttt{DXC$_{5}$} and \texttt{ECC} (ten and one, respectively). According to the results in \figurename~\ref{fig:imagenet_vs_ssl} in the main paper and \tablename~\ref{tab:ssl_vs_imagenet} in Appendix, with  the same number of training epochs, the best self-supervised ImageNet models, such as SimCLR-v1, SeLa-v2, and Barlow Twins, achieve superior performance over supervised ImageNet models in both target tasks.  

\smallskip
\noindent\textbf{3) Supervised ImageNet model vs. domain-adapted models.} We compare the training time of the in-domain pre-trained models to ImageNet counterparts. According to the results in \tablename~\ref{tab:domain_adapted_vs_imagenet_convergence}, (1) ChestX-ray14 and CheXpert models consistently outperform ImageNet models in terms of convergence time in most cases, and (2) The overall trend showcases the faster convergence of domain-adapted pre-trained models (\ie ImageNet$\rightarrow$CheXpert and ImageNet$\rightarrow$ChestX-ray14)  compared to the corresponding ImageNet models.

\section{Self-supervised Learning Methods}
\label{sec:appendix_ssl}
\noindent\textbf{InsDis~\cite{Wu2018insdis}:} 
InsDis treats each image as a distinct class and trains a non-parametric classifier to distinguish between individual classes based on noise-contrastive estimation (NCE)~\cite{gutmann2010noise}. InsDis introduces a feature memory bank maintaining a large number of noise samples (referred to as negative samples), to avoid exhaustive feature computing.

\noindent\textbf{MoCo-v1~\cite{He2020MocoV1} and MoCo-v2~\cite{chen2020improved}:} 
MoCo-v1 creates two views by applying two independent data augmentations to the same image $X$, referred to as positive samples. Like InsDis, the images other than $X$ are defined as negative samples stored in a memory bank. Additionally, a momentum encoder is proposed to ensure the consistency of negative samples as they evolve during training. Intuitively, MoCo-v1 aims to increase the similarity between positive samples while decreasing the similarity between negative samples. 
Through simple modifications inspired by SimCLR-v1~\cite{Chen2020Simple}, such as a non-linear projection head, extra augmentations, cosine decay schedule, and a longer training time to MoCo-v1, MoCo-v2 establishes a stronger baseline while eliminating large training batches.

\noindent\textbf{SimCLR-v1~\cite{Chen2020Simple} and SimCLR-v2~\cite{chen2020big}:} 
SimCLR-v1 is proposed independently following the same intuition as MoCo. However, instead of using special network architectures (\eg a momentum encoder) or a memory bank, SimCLR-v1 is trained in an end-to-end fashion with large batch sizes.  Negative samples are generated within each batch during the training process. In SimCLR-v2, the framework is further optimized by increasing the capacity of the projection head and incorporating the memory mechanism from MoCo to provide more negative samples than SimCLR-v1.

\noindent\textbf{BYOL~\cite{grill2020bootstrap}:} 
Conventional contrastive learning methods such as MoCo and SimCLR relies on a large number of negative samples. As a result, they require either a large memory bank (memory consuming) or a large batch size (computational consuming).
On the contrary, BYOL avoids the use of negative pairs by leveraging two encoders, named online and target, and adding a predictor after the projector in the online encoder. BYOL thus maximizes the agreement between the prediction from the online encoder and the features computed from the target encoder. The target encoder is updated with the momentum mechanism to prevent the collapsing problem.

\noindent\textbf{PIRL~\cite{Misra2020Self}:} 
Instead of using instance discrimination objectives like InsDis and MoCo, PIRL adapts Jigsaw and Rotation as proxy tasks. Specifically, the positive samples are generated by applying Jigsaw shuffling or rotating images by \{$0^{\circ}$, $90^{\circ}$, $180^{\circ}$, $270^{\circ}$\}. PIRL defines a loss function based on noise-contrastive estimation (NCE) and uses a memory bank following InsDis. In this paper, we only benchmark PIRL with Jigsaw shuffling, which yields better performance than its rotation counterpart.

\noindent\textbf{DeepCluster-v2~\cite{caron2021unsupervised}:}
DeepCluster~\cite{caron2018deep} learns features in two phases: (1) self-labeling, where pseudo labels are generated by clustering data points using the prior representation--- yielding cluster indexes for each sample; (2) feature-learning, where the cluster index of each sample is used as a classification target to train a model. The two phases are performed repeatedly until the model converges. Rather than classifying the cluster index, DeepCluster-v2 explicitly minimizes the distance between each sample and the corresponding cluster centroid. DeepCluster-v2 finally applies stronger data augmentation, a MLP projection head, a cosine decay schedule, and multi-cropping to improve the representation learning. 

\noindent\textbf{SeLa-v2~\cite{caron2021unsupervised}:}
Similar to clustering methods, SeLa~\cite{asano2020selflabelling} requires a two-phase training (\ie self-labeling and feature-learning). However, instead of clustering the image instances, SeLa formulates self-labeling as an optimal transport problem, which can be effectively solved by adopting the Sinkhorn-Knopp algorithm. Similar to DeepCluster-v2, the updated SeLa-v2 applies stronger data augmentation, a MLP projection head, a cosine decay schedule, and multi-cropping to improve the representation learning.

\noindent\textbf{PCL-v1 and PCL-v2~\cite{li2021prototypical}:} 
PCL-v1 combines contrastive learning and clustering approaches to encode the semantic structure of the data into the embedding
space. Specifically, PCL-v1 adopts the architecture of MoCo, and incorporates clustering in representation learning. Similar to clustering-based feature learning, PCL-v1 has self-labeling and feature-learning phases. In self-labeling phase, the features obtained from the momentum encoder are clustered, in where each instance is assigned to multiple prototypes (cluster centroids) with different granularity. 
In the feature-learning phase,  PCL-v1 extends the noise-contrastive estimation (NCE) loss to ProtoNCE loss which can push each sample closer to its assigned prototypes. PCL-v2 is developed by applying the aforementioned techniques to promote the representation learning.

\noindent\textbf{SwAV~\cite{caron2021unsupervised}:}
SwAV takes advantages of both contrastive learning and clustering techniques. Similar to SeLa, SwAV calculates cluster assignments (codes) for each data sample with the Sinkhorn-Knopp algorithm. However, SwAV performs online cluster assignments, \ie at the batch level instead of epoch level. Compared with contrastive learning approaches such as MoCo and SimCLR, SwAV ``swapped" predicts the codes obtained from one view using the other view rather than comparing their features directly. Additionally, SwAV proposes a multi-cropping strategy, which can be adopted by other methods to consistently improve their performance.

\noindent\textbf{InfoMin~\cite{tian2020makes}:} 
InfoMin hypothesizes that good views (or positive samples) should only share label information w.r.t the downstream task while throwing away irrelevant factors, which means optimal views for contrastive representation learning are task-dependent. Following this hypothesis, InfoMin optimizes data augmentations by further reducing mutual information between views.

\noindent\textbf{Barlow Twins~\cite{zbontar2021barlow}:} This method consists of two online encoders that are fed by two augmented views of the same image. The model is trained by making the cross-correlation matrix of two encoders' outputs as close to the identity matrix as possible. 
As a result,(1)  the similarity between representations of two views is maximized, which is similar to the ultimate goal of  contrastive learning, and (2) the redundancy between the components of two representations is minimized. 

\end{document}